\pdfoutput=1

\documentclass[11pt]{article}

\usepackage[preprint]{acl}
\usepackage{times}
\usepackage{latexsym}
\usepackage{enumitem}
\usepackage{booktabs}
\usepackage{amsmath}
\usepackage{algorithm}
\usepackage{algpseudocode}

\usepackage[T1]{fontenc}

\usepackage[utf8]{inputenc}
\usepackage{longtable}
\usepackage{microtype}
\usepackage{multirow} 

\usepackage{inconsolata}

\usepackage{graphicx}

\title{METEOR: Evolutionary Journey of Large Language Models  from Guidance to Self-Growth }
\author{Jiawei Li\textsuperscript{1}, Xiaoang Xu\textsuperscript{2}, Yang Gao\textsuperscript{1}\thanks{Corresponding author.}\\
        \textsuperscript{1} School of Computer Science and Technology, Beijing Institute of Technology \\
        \textsuperscript{2} Beijing University of Posts and Telecommunications, School of Artificial Intelligence \\
        \texttt{\{jwli, gyang\}@bit.edu.cn}}

\begin{document}
\maketitle
\begin{abstract}
Model evolution enables learning from feedback to refine experiences and update skills, transforming models from having no domain knowledge to becoming domain experts. However, there is currently no unified and effective method for guiding this evolutionary process. To address this gap, we propose the Meteor method, which includes three training phases: weak-to-strong data distillation, iterative training, and self-evolution strategies. Each phase maximizes the model's inherent domain capabilities, allowing it to autonomously refine its domain knowledge and enhance performance. Experiments demonstrate that our approach significantly improves accuracy, completeness, relevance, coherence, and reliability across domain-specific tasks. Our code can be found at \url{https://github.com/DIRECT-BIT/METEOR}.
\end{abstract}

\section{Introduction}

The development of large language models (LLMs) has ushered in a new era in the field of natural language processing (NLP), showcasing remarkable general capabilities across a wide range of applications~\cite{DBLP:journals/corr/abs-2303-08774,DBLP:journals/corr/abs-2407-10671,DBLP:journals/aiopen/YangSLLLLGH24,DBLP:journals/corr/abs-2403-05530}. However, despite their outstanding performance on general tasks, the training of a highly versatile LLM demands substantial computational resources and financial investment. These high costs restrict their use in many situations, particularly in specific domains. In specific domains, there is often no need for a general purpose intelligent model; instead, a model that acts as an expert within a particular domain is more desirable. Furthermore, these domain-specific expert models should be trained at lower costs and easily deployed in their respective fields. Therefore, finding efficient ways to create a domain-specific expert model has become a key research focus in the development of LLMs~\cite{ling2024domainspecializationkeymake,DBLP:conf/acl/LiYBZLSLSYWLXBF24}.

Some studies leverage the inherent capabilities of LLMs combined with domain-specific external enhancements to enable their application in specific domains. These methods involve explicitly or implicitly acquiring domain knowledge from external knowledge bases~\cite{DBLP:conf/sigir/0001SXMSY23,DBLP:journals/jmlr/IzacardLLHPSDJRG23,DBLP:journals/corr/abs-2301-04589} or utilizing domain-specific tools to assist LLMs in specific domains~\cite{DBLP:journals/bioinformatics/JinYCL24,DBLP:conf/nips/LiHQYLLWQGHZ0LC23,DBLP:journals/corr/abs-2303-16434}. However, these methods rely on the model's strong general capabilities and are typically applicable only to models with a large number of parameters, making them costly to deploy and limiting their widespread adoption. Several researchers have explored model evolution approaches to enhance domain-specific capabilities. \citet{DBLP:journals/corr/abs-2303-17564} achieves evolution in the financial domain through extensive manually annotated domain data, yet this approach proves challenging to scale due to the difficulty in data acquisition. While~\citet{DBLP:journals/corr/abs-2406-04151} proposes utilizing general large models for supervision and feedback on domain model-generated data, this methodology remains constrained by the performance ceiling of the supervising model. Although~\citet{DBLP:journals/tmlr/SinghCAAPGLH0XP24} made breakthrough progress in the coding domain through self-generated data and self-training, eliminating dependence on human annotations and large models, their approach has not yet been effectively extended to other specific domains.


\begin{figure*}[t]
\centering
\includegraphics[width=0.90\linewidth]{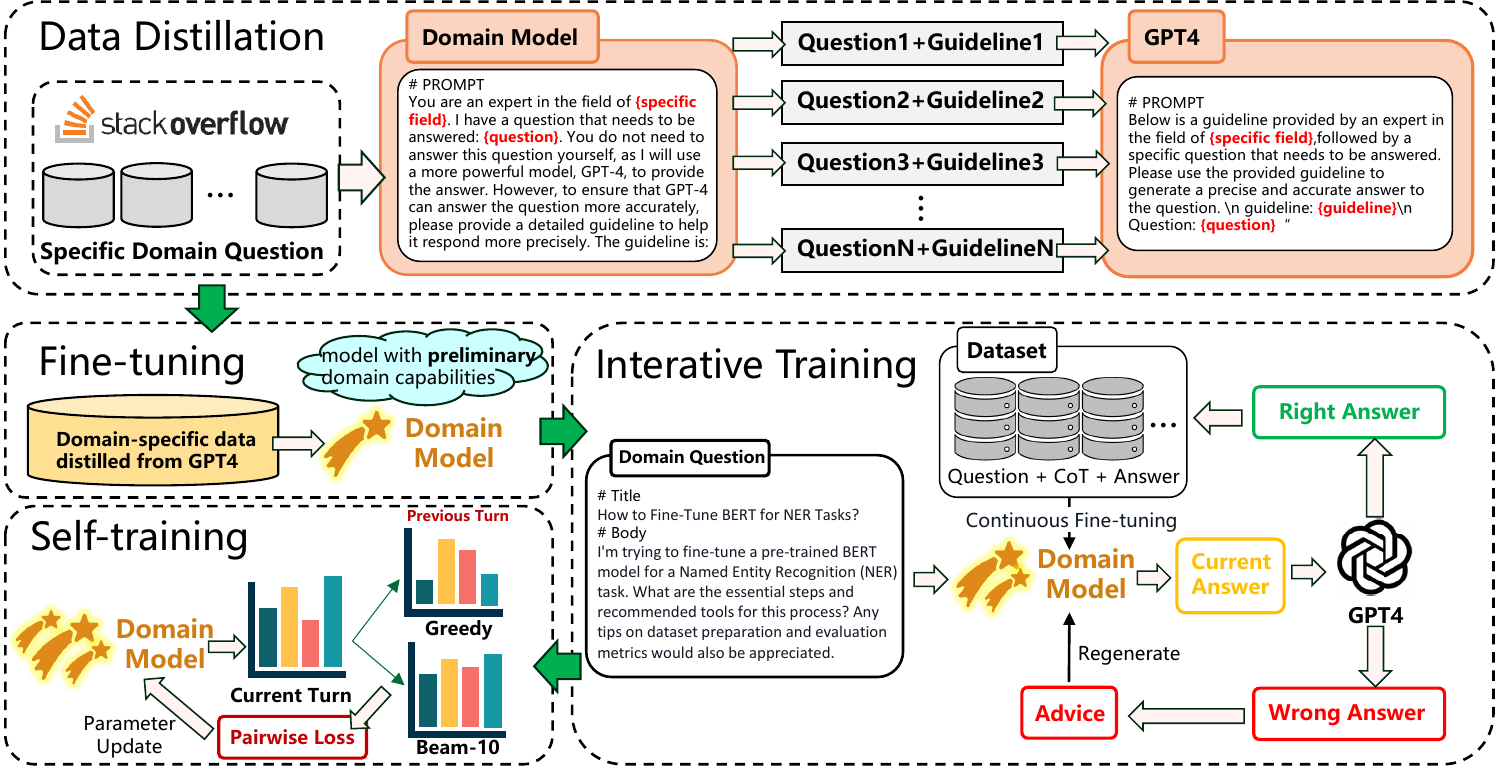} 
\caption{Overview of the METEOR method, which is structured into three phases. Initially, weak-to-strong knowledge distillation is applied, wherein the distilled data is used to train the model, facilitating the initial acquisition of domain-specific capabilities. This is succeeded by iterative training, which further refines the model's domain expertise. Ultimately, self-training is conducted, enabling the model to achieve the proficiency of a domain expert.}
\label{fig:overall}
\end{figure*}

To address the challenges faced by current models in domain-specific applications, we propose a self-evolution method named \textbf{METEOR}, a weak-to-strong evolution framework that enables LLMs to progressively evolve from supervised guidance to autonomous enhancement.
Meteor offers a comprehensive training framework that guides an LLM from having no domain expertise to becoming a domain expert. This framework consists of three key stages: an initial fine-tuning stage, which aims to impart basic domain knowledge to the LLM; an iterative training stage, where the LLM's domain knowledge is further enhanced with the guidance of a strong general LLM (like GPT-4~\cite{DBLP:journals/corr/abs-2303-08774}); and a self-training stage, which aims to further improve the LLM's capabilities. The overall training architecture of Meteor is illustrated in Figure~\ref{fig:overall}.

Knowledge distillation from strong models is an effective approach to address the scarcity of domain-specific data. However, due to the distributional discrepancies between strong and weak models, knowledge distilled directly from the strong model's distribution may not be efficiently utilized by the weak model. To overcome this limitation, we propose a weak-to-strong strategy. In this approach, the weak model first provides guidelines for the domain-specific questions to be distilled, based on its own distribution. The strong model then generates answers to these questions according to the guidelines, thereby aligning the knowledge distribution between the strong and weak models.


Following knowledge distillation, models acquire initial domain capabilities but are still insufficient to independently address complex problems in specific domains. Therefore, further domain evolution is necessary. 
Following knowledge distillation, models acquire initial domain capabilities, enabling them to perform basic domain-specific tasks. However, these capabilities are limited to direct problem-solving without the ability to evaluate and refine their own outputs. This limitation necessitates a guided evolution phase where strong models provide structured feedback, helping domain models develop self-examination abilities through iterative refinement~\cite{DBLP:conf/nips/ShinnCGNY23,DBLP:conf/nips/ZelikmanWMG22}. Once models develop self-examination capabilities through guided reflection, they become capable of identifying potential flaws in their reasoning. This creates the foundation for autonomous improvement. Building on the theoretical relationship between increased inference FLOPs and enhanced model capabilities~\cite{DBLP:journals/corr/abs-2408-03314}, we introduce a self-training approach where models leverage their acquired self-examination abilities to progressively enhance their performance through computational scaling, potentially surpassing their strong supervisors.



The main contributions of this paper are as follows:

\begin{itemize}[leftmargin=*]
\item We introduce a weak-to-strong evolution framework that enables LLMs to progressively evolve from supervised guidance to autonomous enhancement method named METEOR. Experimental results demonstrate that each phase of METEOR enhances the performance of models in specific domain applications.
\item We propose a model data distillation method based on the weak-to-strong strategy, achieving alignment of domain knowledge distribution between strong and weak models.
\item We present an evolution method guided by strong models based on reflection, as well as a self-training evolution method for models.
\end{itemize}

\section{METEOR}

The METEOR method consists of three distinct phases: weak-to-strong data distillation, iterative training, and self-evolution strategies. In each phase, the model fully utilizes its existing capabilities to strengthen its domain expertise. Each subsequent phase builds upon the advancements made in the previous one, employing different techniques to further evolve the model. The following sections will detail the specific methods employed in each phase of METEOR, illustrating how these strategies collectively contribute to the model's enhanced performance and domain-specific knowledge.

\subsection{Weak-to-strong Domain Data Distillation}
\label{sec:weak2strong_sec}

For a LLM lacking domain-specific capabilities, knowledge distillation is proven to be an effective method for injecting domain knowledge~\cite{DBLP:journals/corr/abs-2305-15062,DBLP:conf/emnlp/ZhangCJYCCLWZXW23,DBLP:conf/cikm/ZhangY23}. However, our experiments indicate that directly distilling domain data from a strong general model to train a domain-specific model is not particularly effective. As illustrated in Table~\ref{tab:weak2strong} of Appendix~\ref{sec:appendix1}, there is a discrepancy between the responses of GPT-4 to domain-specific questions and the solutions provided by the weak model. This discrepancy arises from differences in their cognitive distributions, making it challenging to train and learn using directly distilled data.

\begin{figure}[t]
\centering
\includegraphics[width=1\linewidth]{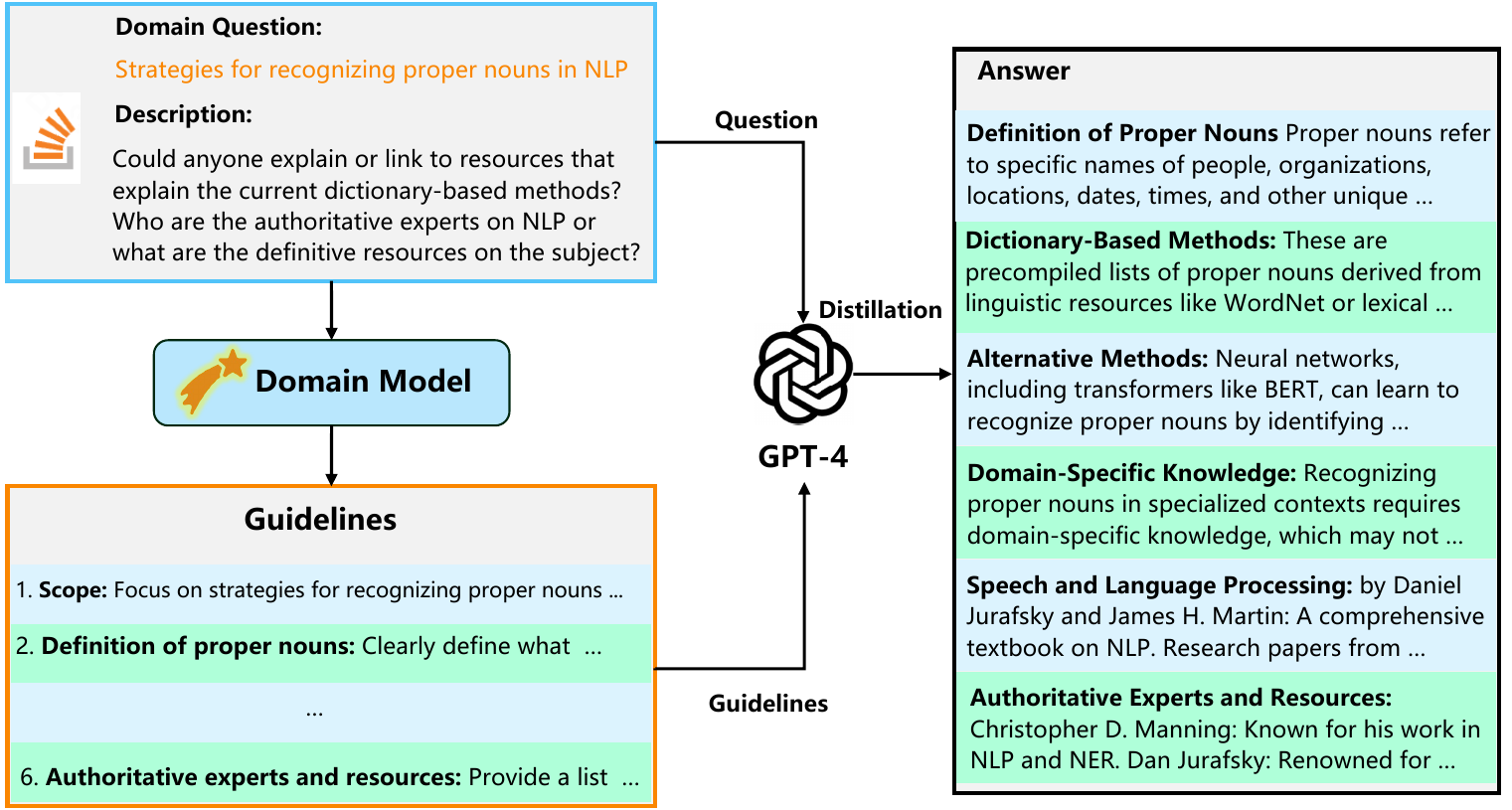} 
\caption{Illustration of the weak-to-strong knowledge distillation process. Initially, a domain question is input into the domain model to obtain a guideline. The strong model then uses this guideline, provided by the weak model, along with the original question, to generate and distill domain-specific data. }
\label{fig:weak2strong_figure}
\end{figure}

To address this issue, we propose a domain data distillation method based on a weak-to-strong strategy~\cite{DBLP:conf/icml/BurnsIKBGACEJLS24}. As illustrated in Figure~\ref{fig:weak2strong_figure}, to align the distribution discrepancies between the strong and weak models, we guide the strong model to distill domain data according to the instructions from the weak model. Specifically, when presented with a domain-specific question, we first input the question into the weak model. Instead of generating an answer, the weak model produces a \textbf{guideline} based on the prompt shown in Figure~\ref{fig:weak2strong_figure}. This guideline outlines the steps the weak model believes should be followed to address the question.

Subsequently, we input the generated guideline along with the domain question into the strong model. The strong model then generates the answer to the domain question by following the guideline's steps or refines the answer based on the guideline.

Through these steps, we obtain a set of data pairs consisting of domain questions and their corresponding answers distilled from the strong model. This distilled data is used to fine-tune the weak model, enabling it to acquire preliminary domain-specific capabilities, thus completing the first stage of the Meteor model's evolution. In Section~\ref{sec:experiment}, we demonstrate that the data obtained using the weak-to-strong distillation strategy is superior to that obtained through direct data distillation, thereby validating the effectiveness of this approach.

\subsection{Model Evolution}

Training models using domain data obtained through knowledge distillation can impart preliminary domain-specific capabilities to the LLMs. However, this approach falls short of elevating the models to the level of true domain experts due to the absence of expert-level reasoning abilities. Therefore, further model evolution methods are required to enhance the domain capabilities of these models.

\begin{figure}[t]
\centering
\includegraphics[width=1\linewidth]{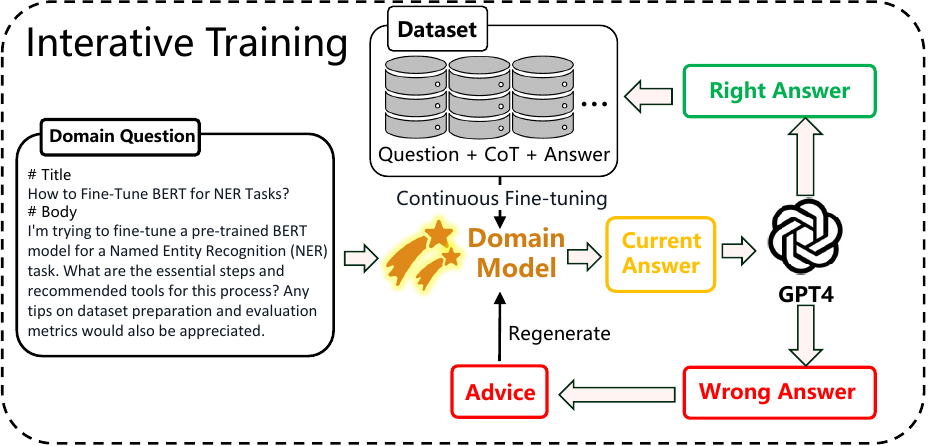} 
\caption{Illustration of he iterative evolution process guided by a strong model. Upon receiving domain-specific data, the model employs CoT reasoning to generate answers and reasoning paths. These are evaluated by GPT-4, which provides confirmation if correct or offers suggestions for refinement if incorrect. This iterative process continues until the answer is validated or the maximum iteration limit is reached.}
\label{fig:fig3}
\end{figure}

The reflection mechanism, which involves analyzing the model's previous reasoning processes and answers to identify errors and provide feedback, has been shown to significantly improve the model's reasoning abilities and performance in downstream applications~\cite{DBLP:conf/nips/ShinnCGNY23,DBLP:conf/nips/MadaanTGHGW0DPY23}. Consequently, we have designed an algorithm for further evolving the model within specific domains, based on the reflection mechanism of LLMs, as illustrated in Figure~\ref{fig:fig3}.

\subsubsection{Data Refinement and Iterative Training for Domain-Specific Models}
\label{sec:interative_sec}

The key aspect of the reflection mechanism is to provide accurate feedback on the model's output. However, models that have been developed through knowledge distillation possess only preliminary domain capabilities and are unable to deliver sufficiently accurate feedback. Therefore, at this stage, we propose using strong models, such as GPT-4~\cite{DBLP:journals/corr/abs-2303-08774}, to provide feedback for the evolution of domain-specific models. The overall algorithm is shown in Algorithm~\ref{alg:iter}.

\begin{algorithm}[t]
\caption{Data Refinement and Iterative Training}
\label{alg:iter}
\begin{algorithmic}[1]
\Require Question set $\mathcal{Q}$, Model $M$, GPT-4, Maximum iterations $N$, Batch size $K$
\Ensure Updated Model $M$
\State $\text{training\_buffer} \gets \{\}$

\For{$Q$ in $\mathcal{Q}$}
    \State $\text{history} \gets \{\}$
    \While{$n < N$}
        \If{$n = 0$}
            \State $R_n, A_n \gets M(Q)$ \Comment{Initial attempt}
        \Else
            \State $R_n, A_n \gets M(Q, \text{guide}_n)$ \Comment{Guided attempt}
        \EndIf
        
        \State $(status, \text{explanation}, \text{guide}) \gets \text{GPT-4}(Q, R_n, A_n)$
        \State $\text{history}.\text{append}(R_n, \text{explanation})$
        
        \If{$status = \text{"CORRECT"}$}
            \State $\text{training\_buffer}.\text{append}(Q, \text{history})$
            \If{$|\text{training\_buffer}| = K$}
                \State $M \gets \text{UpdateModel}(M, \text{training\_buffer})$ \Comment{Fine-tune model}
                \State $\text{training\_buffer} \gets \{\}$ \Comment{Clear buffer}
            \EndIf
            \State \textbf{break}
        \EndIf
        \State $n \gets n + 1$
    \EndWhile
\EndFor
\Return $M$
\end{algorithmic}
\end{algorithm}

The specific method involves inputting a batch of domain data, where the current domain model generates answers and their rationales for the given questions using the Chain-of-Thought (CoT) approach~\cite{DBLP:conf/nips/Wei0SBIXCLZ22}. At this stage, we combine the question, rationale, and answer, which are then input into GPT-4. GPT-4 evaluates whether the rationale and answer are correct. Within the constraints of the context, if GPT-4 deems the answer correct, it returns a correct identifier.
If GPT-4 identifies the answer as incorrect, it provides suggestions for modifying the rationale and offers a guiding sentence to help the model produce the correct rationale. We re-input the domain question, original rationale and answer, along with GPT-4's suggestions and guiding sentence, into the model, which then generates a new answer. This process continues until GPT-4 considers the answer correct or the maximum number of iterations, $N$, is reached.

Once the iteration stops, we combine the domain questions, multiple rounds of rationales, GPT-4's corrections, and guiding sentences from the batch to form the final domain training data. During data construction, we convert the guiding sentences provided by GPT-4 into discourse markers (like Hmm). This approach serves two main purposes: 1. To ensure that the constructed data reflects a strategy where the model independently derives the corrected answer, rather than relying on externally provided guiding sentences for correction. 2. To maintain the coherence of the reasoning statements throughout the process.

After obtaining the domain data, we employ an iterative training approach to train the model. Specifically, when the amount of data in the dataset reaches a multiple of $K$, we fine-tune the model using the current dataset. The fine-tuned model is then reintroduced into the iterative process described above. Throughout the iterative training process, the domain model's capability in the specific domain gradually improves, and the quality of the generated domain data continuously enhances.

\subsubsection{Self-Evolution of Domain Capabilities through Inference Strategy Optimization}
\label{sec:self_train_sec}

Additionally, we aim for the model to evolve its domain capabilities independently, without relying on strong models. To achieve this, we propose a method for the model to autonomously enhance its domain capabilities.

Increasing FLOPs during inference has been shown to effectively enhance model performance in downstream tasks~\cite{DBLP:journals/corr/abs-2408-03314}. Inspired by this concept, we hypothesize that different inference strategies yield varying outcomes. We aim for the model's outputs to closely align with those generated by high-FLOPs strategies, while diverging from those produced by low-FLOPs strategies. Therefore, we employ beam search~\cite{DBLP:conf/aclnmt/FreitagA17} as the high-FLOPs strategy and greedy search as the low-FLOPs strategy, using contrastive learning to develop a self-training method for the model. Experiments in Section~\ref{sec:experiment} demonstrate the effectiveness of this approach.

\begin{table}[t]
\begin{tabular}{lccccc}
\hline
Category & ML   & \multicolumn{1}{l}{DL} & NLP  & CV   & Total \\ \hline
Scale    & 4605 & 2092                    & 1881 & 1698 & 10276 \\ \hline
\end{tabular}
\caption{Total volume of data scraped from Stack Overflow. The categories ML, DL, NLP, and CV represent Machine Learning, Deep Learning, Natural Language Processing, and Computer Vision, respectively. }
\label{tab:table1}
\end{table}

\section{Experimental Setups}
\label{sec:experiment_setups}

\begin{table*}[t]
\begin{tabular}{llcccccc}
\hline
\multicolumn{2}{l}{Model}                    & Accuracy & Completeness & Relevance & Coherence & Reliability & GPT4 Score \\ \hline
\multirow{2}{*}{LLaMA3} & w/o. & 21.3\%  & 27.8\%      & 25.6\%   & 19.9\%   & 17.8\%     & 5.02       \\
                                & Meteor     & \textbf{78.7\%}  & \textbf{72.2\%}     & \textbf{74.4\%}   & \textbf{80.1\%}   & \textbf{82.2\%}     & \textbf{9.17}       \\ \hline
\multirow{2}{*}{Qwen2}       & w/o. & 31.6\%  & 36.5\%      & 39.7\%   & 36.5\%   & 32.9\%     & 6.88       \\
                                & Meteor     & \textbf{68.4\%}  & \textbf{63.5\%}      & \textbf{60.3\%}   & \textbf{63.5\%}   & \textbf{c\%}     & \textbf{9.28}       \\ \hline
\end{tabular}
\caption{The overall results comparing the quality of domain question responses by LLaMA3 and Qwen2, both with and without Meteor training. Here, LLaMA refers to LLaMA3-8B-Chat, Qwen2 refers to Qwen2-7B-Instruct, and w/o. denotes models that have not undergone Meteor training. }
\label{tab:over_all}
\end{table*}

\paragraph{Datasets.} We use the field of advanced computer education as the specific domain to validate the effectiveness of the proposed Meteor method. To obtain high-quality domain data, we scraped data from Stack Overflow across four categories: Machine Learning (ML), Deep Learning (DL), Natural Language Processing (NLP), and Computer Vision (CV), totaling 10,276 entries. The data distribution across these categories is shown in Table~\ref{tab:table1}. From this dataset, we randomly selected 1000 entries as test data, with the remaining data used for training.

\paragraph{Metrics.} Following the evaluated method proposed by~\citet{DBLP:conf/nips/ZhengC00WZL0LXZ23}, we use GPT-4 as a judge to evaluate both data quality and model performance. When comparing the distilled data quality with and without the use of guidelines, GPT-4 is used to score the data, where higher scores indicate better quality. In evaluating the domain-specific answers generated by the model, GPT-4 provides scores based on five criteria: accuracy, completeness, relevance, coherence, and reliability, allowing for a comprehensive assessment of the model's domain capabilities. The prompt for GPT-4's pairwise comparison across five evaluation criteria is presented in Table~\ref{tab:pairwise} of Appendix~\ref{sec:app_prompts}.

\paragraph{Model and parameters setting.} We selected LLaMA3-8B-Chat~\cite{DBLP:journals/corr/abs-2407-21783} and Qwen2-7B-Instruct~\cite{DBLP:journals/corr/abs-2407-10671} as the baseline models for domain model training in this study. For GPT-4, we utilized the publicly available API model, GPT-4o. During the iterative evolution of the model, we set the maximum number of iterations to $N=3$, with a cumulative data amount of $K=10$ during each iteration. The beam size for beam search was set to 10, and the maximum token length for the domain evolution model was configured to be 4096.

\section{Experimental Results}
\label{sec:experiment}

\subsection{Overall Results}
We compared the performance changes across various dimensions before and after applying the Meteor method for domain capability evolution of LLMs. For the accuracy, completeness, relevance, coherence, and reliability, we generated responses to test set questions using both the Meteor-trained LLMs and the non-Meteor-trained LLMs. GPT-4 was then used to determine which model's responses performed better in each of these dimensions. For the GPT-4 Score, GPT-4 directly evaluated the answers generated by the LLMs before and after evolution, and the average score was calculated.

As shown in Table~\ref{tab:over_all}, after the Meteor evolution, LLaMA3-8B-Chat and Qwen2-7B-Instruct achieved improvements of 78.66\%, 72.13\%, 74.34\%, 80.10\%, 82.13\%, and 68.36\%, 63.43\%, 60.24\%, 63.43\%, 67.09\% respectively in accuracy, completeness, relevance, coherence, and reliability. Additionally, the GPT-4 Score after evolution was significantly higher than before, demonstrating the effectiveness of the Meteor method.

\subsection{Effectiveness of Weak-to-Strong Data Distillation}
In Section~\ref{sec:weak2strong_sec}, we introduced a domain knowledge distillation method based on a weak-to-strong strategy. To evaluate the effectiveness of this approach, we used GPT-4 to score the distilled data. By comparing the GPT-4 Scores of data distilled with and without guidelines, we demonstrate the effectiveness of the weak-to-strong strategy in enhancing the quality of distilled data. The specific results are shown in Table~\ref{tab:guideline_score}.

As shown in Table~\ref{tab:guideline_score}, across various domains, the data distilled with guidelines achieved significantly higher GPT-4 scores than the data distilled without guidelines. Specifically, the GPT-4 scores increased by 3.29, 3.27, 3.34, and 3.32 points in ML, DL, NLP, and CV, respectively, indicating that the quality of domain data distilled with guidelines is superior.

\begin{table}[h]
\begin{tabular}{lcc}
\hline
\multirow{2}{*}{Category} & \multicolumn{2}{c}{GPT-4 Score}  \\ \cline{2-3} 
                          & with guidelines & w/o guidelines \\ \hline
ML                        & \textbf{9.09}            & 5.80           \\
DL                        & \textbf{9.07}            & 5.80           \\
NLP                       & \textbf{9.05}            & 5.71           \\
CV                        & \textbf{9.05}            & 5.73           \\ \hline
Avg.                      & \textbf{9.07}            & 5.77           \\ \hline
\end{tabular}
\caption{This table illustrating the quality scores assigned by GPT-4 to data distilled with and without guidelines. The results indicate that data distilled using guidelines consistently receive higher scores across all domains. }
\label{tab:guideline_score}
\end{table}

Additionally, we assessed whether data distilled with guidelines can more effectively enhance the model's domain capabilities. We fine-tuned the model using two sets of data, one with guidelines and one without, and evaluated the GPT-4 Scores of the answers generated by the models on the test set. As shown in Table~\ref{tab:guideline_score}, models fine-tuned with data distilled using guidelines scored higher across all aspects compared to those fine-tuned with data distilled without guidelines, further confirming the effectiveness of the weak-to-strong data distillation strategy.

\begin{table}[h]
\begin{tabular}{lcc}
\hline
\multirow{2}{*}{Model} & \multicolumn{2}{c}{GPT-4 Score}  \\ \cline{2-3} 
                       & with guidelines & w/o guidelines \\ \hline
LLaMA3         & \textbf{9.04}            & 5.38           \\
Qwen2      & \textbf{9.22}            & 6.13           \\ \hline
\end{tabular}
\caption{This table presenting GPT-4 scores for answers generated by models fine-tuned with distilled data. The results demonstrate that models fine-tuned with data distilled using guidelines exhibit superior performance on domain-specific tasks. }
\label{tab:guideline_finetune}
\end{table}

\subsection{Effectiveness of Iterative Training and Data Refinement}

In Section~\ref{sec:interative_sec}, we introduced the Data Refinement and Iterative Training for Domain-Specific Models. o validate the effectiveness of this approach, we employed GPT-4 to evaluate the accuracy, completeness, relevance, coherence, and reliability of the answers generated by the models before and after evolution. The results are presented in Table~\ref{tab:iterative}.

\begin{table}[h]
\begin{tabular}{lcccc}
\hline
\multirow{2}{*}{} & \multicolumn{2}{c}{LLaMA3} & \multicolumn{2}{c}{Qwen2} \\ \cline{2-5} 
                  & w.               & w/o.    & w.               & w/o.   \\ \hline
Acc.          & \textbf{72.7\%}  & 27.2\%  & \textbf{54.9\%}  & 45.1\% \\
Com.      & \textbf{63.1\%}  & 36.9\%  & \textbf{53.7\%}  & 46.3\% \\
Rele.         & \textbf{65.4\%}  & 34.6\%  & \textbf{51.3\%}  & 48.6\% \\
Coh.         & \textbf{72.3\%}  & 22.7\%  & \textbf{57.6\%}  & 42.4\% \\
Reli.       & \textbf{62.9\%}  & 37.1\%  & \textbf{60.8\%}  & 39.2\% \\ \hline
\end{tabular}
\caption{The effectiveness of Data Refinement and Iterative Training for Domain-Specific Models. Acc., Com., Rele., Coh., and Reli. represent Accuracy, Completeness, Relevance, Coherence, and Reliability, respectively. LLaMA3 represents LLaMA3-8B-Chat Model. Qwen2 represents Qwen2-7B-Instruct Model. 'w.' denotes models that have undergone iterative self-evolution training, while 'w/o.' denotes models that have not.}
\label{tab:iterative}
\end{table}

The results in Table~\ref{tab:iterative} indicate that both LLaMA3-8B-Chat~\cite{DBLP:journals/corr/abs-2407-21783} and Qwen2-7B-Instruct~\cite{DBLP:journals/corr/abs-2407-10671} showed improved performance across all metrics after evolution guided by GPT-4, demonstrating the effectiveness and generalizability of this method.

\subsection{Effectiveness of Self-Evolution Method}

In Section~\ref{sec:self_train_sec}, we proposed that model evolution should be independent of strong model constraints and introduced a self-evolution method based on varying FLOPs. Table~\ref{tab:self-train} presents the differences in performance across various dimensions, as evaluated by GPT-4, before and after the models self-evolution.

\begin{table}[h]
\begin{tabular}{lcccc}
\hline
\multirow{2}{*}{} & \multicolumn{2}{c}{LLaMa3}        & \multicolumn{2}{c}{QWen2} \\ \cline{2-5} 
                  & w.              & w/o.            & w.               & w/o.   \\ \hline
Acc.          & \textbf{53.8\%} & 46.2\%          & \textbf{54.7\%}  & 45.3\% \\
Com.      & 47.9\%          & \textbf{52.1\%} & \textbf{51.3\%}  & 48.7\% \\
Rele.         & \textbf{51.7\%} & 48.3\%          & \textbf{50.8\%}  & 49.2\% \\
Coh.         & \textbf{53.3\%} & 46.7\%          & \textbf{52.8\%}  & 47.2\% \\
Reli.       & \textbf{52.9\%} & 47.1\%          & \textbf{53.1\%}  & 46.9\% \\ \hline
\end{tabular}
\caption{The effiectiveness of Self-evolution method. Acc., Com., Rele., Coh., and Reli. represent Accuracy, Completeness, Relevance, Coherence, and Reliability, respectively. LLaMA3 represents LLaMA3-8B-Chat Model. Qwen2 represents Qwen2-7B-Instruct Model. 'w.' denotes models that have self-evolution training, while 'w/o.' denotes models that have not.}
\label{tab:self-train}
\end{table}

The results in Table~\ref{tab:self-train} indicate that both LLaMA3-8B-Chat~\cite{DBLP:journals/corr/abs-2407-21783} and Qwen2-7B-Instruct~\cite{DBLP:journals/corr/abs-2407-10671} showed improved performance across all metrics following self-evolution, demonstrating the effectiveness and generalizability of this method.

\section{Related Works}

\subsection{Domain-specific Knowledge Distillation}
In the context of LLMs, knowledge distillation has become a crucial method for enhancing domain-specific knowledge. In specialized fields such as law and medicine, the complexity of terminology often poses challenges~\cite{DBLP:journals/corr/abs-2312-03718,DBLP:conf/emnlp/0001YMWRCYR23}. As a result, the distillation process typically begins with pre-training on domain-specific corpora, followed by fine-tuning using LLMs to construct enhanced data~\cite{DBLP:journals/corr/abs-2305-15062,DBLP:journals/corr/abs-2406-04614}. In the financial fields, distilled data must maintain diversity and complexity to ensure that LLMs can handle varied and intricate scenarios~\cite{DBLP:conf/cikm/ZhangY23,DBLP:journals/corr/abs-2310-04793}. In scientific fields, the distillation of knowledge from long texts and reasoning data is essential~\cite{DBLP:conf/aaai/AbdineCBV24,DBLP:journals/corr/abs-2401-07950,DBLP:journals/corr/abs-2401-00434}. In the education fields, researchers often emphasize knowledge distillation through pre-training on educational corpora and fine-tuning with custom instructions to activate education-specific functions such as open-ended questioning, essay evaluation, and emotional support~\cite{DBLP:journals/corr/abs-2308-02773,DBLP:conf/iclr/Guo0LYCBPHCZSZZ24}. However, few studies focus on whether the inputs distilled from strong models can be efficiently utilized. This paper achieves more efficient domain knowledge distillation by aligning the knowledge distribution between strong and weak models.

\subsection{Self-evolution of LLMs}
Self-evolution of LLMs refers to the capability of LLMs to autonomously acquire, refine, and learn from their own generated experiences~\cite{DBLP:journals/corr/abs-2404-14387}.  Initially, model evolution requires experience acquisition. Self-Align~\cite{DBLP:conf/iclr/LiYZSLZWL24} generates experiences guided by knowledge from over 20 domains to ensure diversity in evolution. In contrast, Self-Instruct~\cite{DBLP:conf/acl/WangKMLSKH23} proposes a self-evolution method without the need for external knowledge guidance. Subsequently, experience refinement is necessary. Self-Talk~\cite{DBLP:conf/acl/UlmerMLSGZ24} ensures data quality by measuring the number of completed sub-goals, while Self-verification~\cite{DBLP:conf/emnlp/WengZX0HLSLZ23} selects data by evaluating the consistency between predicted values and original conditions. Once high-quality experiences are acquired, models need to be updated to enhance performance. ReST~\cite{DBLP:journals/corr/abs-2312-10003} updates LLMs iteratively by blending original training data with newly generated data. FuseLLM~\cite{DBLP:conf/iclr/WanH0QB024} transfers knowledge to the target LLM using probability distributions generated by the source LLM. Finally, the evolved models require evaluation. ChatEval~\cite{DBLP:conf/iclr/ChanCSYXZF024} explores the strengths and weaknesses of model outputs through a debate mechanism. LLM-as-a-judge~\cite{DBLP:conf/nips/ZhengC00WZL0LXZ23} employs large models to evaluate other large models, demonstrating that LLMs can match human judgment and achieve efficient performance evaluation.

\section{Conclusion}
In this study, we explore innovative strategies for enhancing domain-specific capabilities of LLMs through knowledge distillation and self-evolution. We propose the Meteor method, which leverages weak-to-strong data distillation, iterative training, and self-evolution to improve model performance across various metrics. Our approach is validated using data from advanced computer education, with experiments demonstrating significant improvements in accuracy, completeness, relevance, coherence, and reliability. By aligning the knowledge distribution between strong and weak models, we achieve more efficient domain knowledge distillation. Furthermore, our self-evolution method enables models to autonomously refine their capabilities without relying on external models, ensuring adaptability and robustness.

\section{Limitations and Future Works}

\subsection{Limitations}

Despite the significant advances achieved in domain-specific model evolution, several limitations warrant further investigation and improvement in future research.

The primary limitation lies in our approach to validating the distributional discrepancy between strong and weak models. Although we have successfully demonstrated performance improvements through weak-to-strong knowledge distillation, thereby indirectly supporting our hypothesis of domain-specific distributional differences, we lack direct experimental evidence to quantify and visualize these differences. This indirect validation methodology may not fully capture the specific characteristics and extent of the distributional disparities, thus limiting our comprehensive understanding of model behaviors.

Furthermore, the proposed self-evolution methodology demonstrates room for improvement in practical applications. While designed to reduce dependence on strong models, current experimental results indicate limited performance gains. This limitation suggests the necessity for further optimization of our self-evolution strategies to achieve more substantial improvements.

\subsection{Future Works}

Building upon these limitations, we propose several crucial directions for future research:

Firstly, we plan to conduct more rigorous analysis through neural probing techniques to directly quantify the distributional differences between strong and weak models. Specifically, we will track and compare the internal neural activation patterns and token distribution characteristics of both model types when processing domain-specific tasks. This approach will provide more direct experimental evidence, facilitating better understanding and validation of our distributional difference hypothesis.

Secondly, we are committed to developing more efficient self-evolution methodologies. Current research suggests the possibility of surpassing the performance ceiling of strong models, motivating us to explore more advanced self-evolution strategies. Our objective is to design methods capable of continuous performance enhancement, ultimately achieving autonomous evolution that surpasses strong model guidance.

Finally, considering the generic nature of the Meteor approach, we intend to expand its application scope. This includes validating its effectiveness across a more diverse range of domain-specific scenarios and exploring compatibility with various foundation models. Through this extensibility study, we aim to further validate the universality of the Meteor approach and explore its potential in broader application contexts.

\bibliography{mypaper}

\appendix

\section{Example Appendix}
\label{sec:appendix1}

\begin{table*}
\centering
\scalebox{0.6}{
\begin{tabular}{|p{0.9\linewidth}|}
\hline
\textbf{Domain Question} \\
Qustion: Strategies for recognizing proper nouns in NLP 

Describtion: I'm interested in learning more about Natural Language Processing (NLP) and am curious if there are currently any strategies for recognizing proper nouns in a text that aren't based on dictionary recognition? Also, could anyone explain or link to resources that explain the current dictionary-based methods? Who are the authoritative experts on NLP or what are the definitive resources on the subject? \\
\hline
\textbf{GPT-4 Answer} \\
Recognizing proper nouns in Natural Language Processing (NLP) is a critical task for many applications such as information retrieval, machine translation, and named entity recognition. Here are some strategies and resources for recognizing proper nouns, both dictionary-based and non-dictionary-based:

Non-Dictionary-Based Methods
Machine Learning Approaches:

Supervised Learning: Techniques like Conditional Random Fields (CRFs) and Hidden Markov Models (HMMs) are commonly used. These models learn to recognize patterns in labeled training data to identify proper nouns.
Neural Networks: Deep learning models, such as Recurrent Neural Networks (RNNs) and Transformers (e.g., BERT), can be trained on large corpora to recognize proper nouns based on context and syntactic patterns.
Rule-Based Systems:

Pattern Matching: Use regular expressions and linguistic rules to identify proper nouns. For instance, capitalized words following certain parts of speech can be flagged as proper nouns.
Heuristics: Implement heuristics based on punctuation, capitalization, and position in a sentence to detect proper nouns.
Statistical Methods:

Part-of-Speech Tagging: Use statistical models to tag words with their parts of speech, including proper nouns. These models often rely on large annotated corpora for training.
Dictionary-Based Methods
Lexicon-Based Approaches: These methods rely on predefined dictionaries or gazetteers of known proper nouns. The text is scanned to match words against entries in these dictionaries.
Hybrid Methods: Combine dictionary-based recognition with machine learning to improve accuracy. For example, a system might use a dictionary to recognize common proper nouns and a machine learning model to identify less common ones.
Resources and Experts
Books and Papers:

"Speech and Language Processing" by Daniel Jurafsky and James H. Martin is a comprehensive textbook that covers many aspects of NLP, including named entity recognition.
Research papers on named entity recognition (NER) often discuss methods for recognizing proper nouns.
Online Courses and Tutorials:

Coursera and edX offer courses on NLP that cover both traditional and modern techniques.
Online tutorials and documentation for NLP libraries like NLTK, SpaCy, and Stanford NLP.
Experts in the Field:

Christopher Manning: Known for his work in computational linguistics and NLP.
Andrew Ng: While more broadly focused on AI, he has contributed significantly to machine learning applications in NLP.
Conferences and Journals:

Conferences like ACL (Association for Computational Linguistics) and EMNLP (Empirical Methods in Natural Language Processing) are key venues for the latest research in NLP.
Journals such as "Computational Linguistics" and "Journal of Artificial Intelligence Research" publish cutting-edge NLP research.
These strategies and resources should provide a solid foundation for understanding and implementing proper noun recognition in NLP systems.\\
\hline

\textbf{Guidelines} \\

1. **Scope:** Focus on strategies for recognizing proper nouns in natural language text, excluding dictionary-based methods. Discuss the limitations and challenges of dictionary-based approaches and highlight alternative methods that have been proposed or implemented.
2. **Definition of proper nouns:** Clearly define what constitutes a proper noun, including names of people, organizations, locations, dates, times, and other specific entities. Explain the importance of recognizing proper nouns in NLP applications.
3. **Dictionary-based methods:** Provide an overview of current dictionary-based methods for recognizing proper nouns, including:
	* Word lists: The use of precompiled lists of proper nouns, often based on linguistic resources like WordNet or other lexical databases.
	* Gazetteers: Geospatial databases that contain information about geographic locations, often used for proper noun recognition.
	* Entity recognition: The use of machine learning models to identify entities in text, including proper nouns.
4. **Alternative methods:** Discuss alternative strategies for recognizing proper nouns that aren't based on dictionary recognition, including:
	* Machine learning approaches: Explain how machine learning algorithms, such as neural networks or decision trees, can be trained to recognize proper nouns based on patterns in text data.
	* Contextual analysis: Describe how contextual analysis, including sentiment analysis, topic modeling, or entity co-occurrence, can be used to identify proper nouns.
	* Graph-based methods: Explain how graph-based approaches, such as graph neural networks or graph-based entity recognition, can be used to recognize proper nouns.
5. **Challenges and limitations:** Discuss the challenges and limitations of recognizing proper nouns, including:
	* Ambiguity: The difficulty of distinguishing between proper nouns and common nouns.
	* Out-of-vocabulary words: The challenge of recognizing proper nouns that are not present in dictionaries or training data.
	* Domain-specific knowledge: The need for domain-specific knowledge to recognize proper nouns in specific contexts.
6. **Authoritative experts and resources:** Provide a list of authoritative experts in the field of NLP, including researchers and practitioners who have made significant contributions to the development of proper noun recognition techniques. Also, include a list of definitive resources on the subject, such as research papers, books, or online courses. \\

\bottomrule

\end{tabular}

}
\caption{A weak-to-strong example.}
\label{tab:weak2strong}
\end{table*}

\section{Prompts}
\label{sec:app_prompts}

\begin{table*}
\centering
\begin{tabular}{|p{0.9\linewidth}|}
\toprule
\textbf{GPT-4's pairwise comparison prompt:} \\
\hline
Domain Question: [Question]

Answer A (Model 1): 
[First model's answer]

Answer B (Model 2): 
[Second model's answer]

Please compare these two answers and determine which one performs better in each of the following criteria. For each criterion, select the better answer (A or B) and briefly explain why.

1. Accuracy: Which answer provides more accurate domain-specific information? \\
Selection: [A/B] \\
Reason:

2. Completeness: Which answer covers the essential aspects more thoroughly? \\
Selection: [A/B] \\
Reason:

3. Relevance: Which answer better addresses the domain-specific requirements? \\
Selection: [A/B]\\
Reason:

4. Coherence: Which answer presents information in a more logical and organized manner? \\
Selection: [A/B] \\
Reason:

5. Reliability: Which answer demonstrates better technical validity and trustworthiness? \\
Selection: [A/B] \\
Reason:

Overall Winner: [A/B] \\
Key Differences: [Brief summary of main differences between the two answers] \\
\bottomrule
\end{tabular}
\caption{Prompt for GPT-4's pairwise comparison of domain-specific model outputs across five evaluation criteria.}
\label{tab:pairwise}
\end{table*}

\end{document}